\documentclass[runningheads,a4paper]{llncs}

\usepackage{epsfig}
\usepackage{amsmath}
\usepackage{amssymb}
\usepackage{url}
\usepackage{wrapfig}

\begin{document}

\title{A Data Management Approach for Dataset Selection Using Human Computation }

\titlerunning{Dataset Selection Using Human Computation}

\author{Alexandros Ntoulas\inst{1}%
\thanks{Work done while at Microsoft}%
\and Omar Alonso\inst{2}\and Vasileios Kandylas\inst{2}}
\authorrunning{Dataset Selection Using Human Computation}

\institute{Zynga, San Francisco, CA, USA,\\
\email{ntoulas@gmail.com}
\and
Microsoft,
Mountain View, CA, USA, \\
\email{\{omar.alonso, vakandyl\}@microsoft.com}
}

\maketitle

\begin{abstract}
As the number of applications that use machine learning algorithms increases, the need for labeled data useful for training such algorithms intensifies. 

Getting labels typically involves employing humans to do the annotation, which directly translates to training and working costs. Crowdsourcing platforms have made labeling cheaper and faster, but they still involve significant costs, especially for the cases where the potential set of candidate data to be labeled is large. 
In this paper we describe a methodology and a prototype system aiming at addressing this challenge for Web-scale problems in an industrial setting. We discuss ideas on how to efficiently select the data to use for training of machine learning algorithms in an attempt to reduce cost. We show results achieving good performance with reduced cost by carefully selecting which instances to label. Our proposed algorithm is presented as part of a framework for managing and generating training datasets, which includes, among other components, a human computation element.
\end{abstract}

\section{Introduction}
\label{section:introduction}

Several large-scale problems of the Web require readily available labeled data for training machine learning (ML) algorithms. Such problems are, for example, the learning of a ranking function, spam classification, scheduling of page downloads etc. The labeling of the data is usually done by (skilled) humans who examine the data and perform the labeling task according to a predefined set of instructions. 

The recent advent of crowdsourcing has made things more streamlined, but several challenges still remain. One key challenge is to decide which data items to present to the humans for labeling. Given its large scale, the Web is abundant with unlabeled data. However, labeling is a resource-intensive process as the humans have a finite amount of time to work in a day and there is usually some monetary cost associated with the labeling. Given these constraints, we need a way to determine which subset of all the potential data instances to label, so that we can reduce the time and effort of getting a good training set.

Getting labels through human annotation involves performing ``work'' which has an associated cost. The cost has two aspects. First, there is the cost per label, that is, how much money we pay a worker to complete a task and how much time is spent for labeling.\footnote{One can also consider as cost the effect of potentially wrong labels on the machine learned model, but for simplicity here we will consider an average cost per label.} Ideally, this cost should be reduced as much as possible without affecting the generated data set with respect to the training of the ML model. Assuming a fixed monetary and temporal cost per label, we would then aim to get the smallest number of labels that is necessary to generate a ``good'' training set, i.e. a set with which we can train an ML model and achieve performance (classification, ranking, etc.) that is the same as (or comparable to) what we would attain if we had labeled and used for training all the available data. The second aspect of cost is the productivity loss for a scientist or developer who is waiting for the results of a large dataset so he/she can train whatever ML model is required. In some cases the goal is to train an ML model with good performance and complete the training as soon as possible. How many data instances will be labeled or which ones will be used is usually of less importance as the problem domain may be very broad and a quick ``good enough'' solution may be preferable. Assuming again that each label requires a fixed amount of time, we could reduce the overall time and hence the productivity cost either by using more human annotators (but this would increase the monetary cost), or by labeling less data. Therefore, one could reduce both cost aspects by selecting data for labeling in a smart way that will not hurt the quality of the resulting training set.

There are two important questions that we address in this paper. First we need to decide which data items to label. Assuming that we have an abundance of candidate data that we could potentially label, how can we find those that are good for the ML algorithm? The second question is how many data items to label? If we label too few data items we run the risk of having our algorithm under-trained and thus achieving poor accuracy. On the other hand, if we pick too many data items to train on, we might be wasting our resources due to diminishing returns. It might even be the case that some additional data instances do not help at all in improving the accuracy of the model.

We make three contributions with this paper:
\begin{itemize}
\item First, we propose a data management framework for collecting, labeling and evaluating datasets used with ML algorithms. Our framework gives a clear view of the process, identifies the specific steps and provides a straightforward way to replace and test alternatives for the individual components of the process. 
\item Second, we present a new data selection algorithm that uses the concept of feature coverage to select data for labeling. We show that the algorithm generates datasets for training an ML model, which then achieves comparable performance as when trained with all the data. Achieving similar performance with only a subset of the data has important benefits with respect to cost and how soon a trained model can be available for use. 
\item Third, we examine the use of human computation for labeling the data, viewed as an integral aspect of the whole dataset preparation process. The quality of work is treated as an important component of the framework. This allows one to focus on the different ways that their work can affect the end result, or how to control facets of their performance, such as reducing the number of errors or identifying spammers.
\end{itemize}

The rest of the paper is organized as follows. In the next section we discuss the framework for preparing data sets for training. In section \ref{section:algorithm} we present our data selection algorithm. Section \ref{section:evaluation} contains the results of the evaluation of our algorithm compared to two commonly used techniques. We conclude with a discussion of related and future work in sections \ref{section:related} and \ref{section:conclusion} respectively.

\section{Framework and Architecture}
\label{section:framework}

We define our framework in the context of running repeatable tasks that require a lot of labels, a typical industrial setting. In that sense, the \emph{continuous evaluation} of the quality of the ML algorithm is the driving force of this data management approach. Before we describe the architecture, we introduce the following definitions and notation. 

The problem we are attempting to solve can be formalized as follows:\\
Given
\begin{itemize}
\item a (potentially very large) set of unlabeled items $D_u$,
\item a budget $R$,
\item a pool of workers $H$ that can perform work with an incentive $I$,
\item a time period $T$ to complete the task,
\item the cost $c_i$ to label each item $d_i$ within $D_u$.
\end{itemize}
Find a set $D_h$ of $k$ items to label, such that:
\begin{itemize}
\item the ML algorithm achieves good performance,
\item the cost does not exceed the budget $R$: $\sum{c_k} \le R$.
\end{itemize}

We should note that the cost $c_i$ to label each item $d_i$ can be the same for all items or it can be different. For example, we might use a fixed number of workers per item, in which case the cost will be the same for all items. In many cases however multiple labels are collected for each data item and the final label is selected to be the majority or average of all the collected labels for that item. This is done to avoid possible errors made by an individual worker. In yet other cases, if the task is inherently difficult (for example, rating the usefulness of product reviews, or ranking of results) we might use a variable number of workers. If the work item is easy, workers will agree in their labeling and we can stop collecting labels for that item. If the item is hard, there will be disagreements among the collected labels and we may opt to collect more labels until the average is stable or a clear majority emerges. In this latter case the cost per item will not be the same for all items.

\begin{figure}[tpb]
  \centering
  \includegraphics[width=.8\columnwidth]{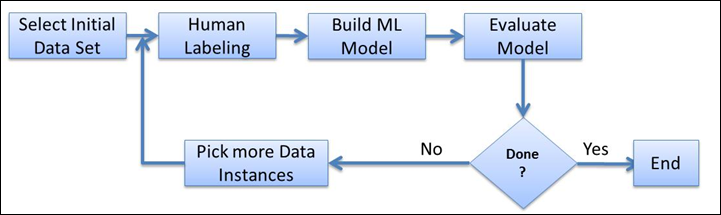}
  \caption{The framework for generating training datasets.}
  \label{fig1}
\end{figure}

The architecture of our framework is shown in Figure~\ref{fig1}. The system is overall composed of six components. We discuss each one of them in more detail next.

\textbf{Select Initial Data Set}. This is the first component and is essentially used to bootstrap the system. There is a variety of different ways that we can use to select the initial dataset for bootstrapping. One way would be to simply give an empty set to the system, so that it decides by itself all of the items that it should label. A different approach would be to take a random sample from the existing data. This is one of the current popular techniques. A third approach would be to take into account budgetary constraints and pick the items more intelligently. This is the focus of our algorithm and we will discuss our approach in more detail later. The algorithm used in this component can be the same as or different from the algorithm used in the \textbf{Pick more data instances} component described later.

\textbf{Human Labeling}.  After we have some data we need to recruit workers so they can perform the work required. We use the term \emph{labeling} to denote the answers provided by the workers for a certain task (or experiment). Labeling can be as simple as a ``yes/no" answer or more complex
requiring pre-qualification of workers and can be performed in a variety of different ways. For example, we can potentially hire people for a particular task and connect them to our system directly (e.g. through a specialized web page, or through an email-based system). This \emph{hired editor} approach is usually expensive as it involves hiring and training people and infrastructure setup. A new and cheaper alternative is to leverage workers via a crowdsourcing model,
where we can connect our system directly to an existing crowdsourcing platform. 

\textbf{Build ML Model}. The particular ML model depends on the problem at hand. For example, certain problems may require a classifier to be employed, while others may require a clustering technique or a technique based on neural-networks. Our system is able to work with any ML model since we will be using a uniform formatting of the data. The goal is to have a toolkit of several ML algorithms and let the user (or even the system automatically) pick and choose which ones to employ for a particular problem.

\textbf{Evaluate ML Model}. A ML model can be evaluated in a variety of different ways. Our system supports the two most common ones, namely cross-validation and the use of a separate test set created specifically for the problem at hand and which is used to evaluate the ML model. Both of these techniques are commonly used for evaluation, but other techniques can also be plugged into this component as needed.

\textbf{Stopping Condition}. The system needs to automatically decide when to stop asking for more data. This can be done in a variety of different ways. For example, the system may decide to stop when we have used up all of our budget or our available time. Another option would be to stop when the ML model has reached a given, predefined goal (for example 90\% accuracy in the classifier).

\textbf{Pick more data instances}. This module selects additional data for labeling. It can be very conservative and only select a few data at a time for labeling. In this case we will do more loops within the system but we will pick the data more ``carefully'' as we will use the output of the ML algorithm more often as guidance for selecting the appropriate instances. Alternatively, the system can be more aggressive and in each round select larger chunks of data to be labeled and sent through the whole work flow again.
 
The algorithm used in this module can be the same as the one used for the initial selection of data. In that case the algorithm would use the same technique to get more instances for labeling, regardless of how the ML model is doing. It is possible however to use a more sophisticated algorithm that takes the output (accuracy) of the ML model into account. For example, the algorithm may decide to include more data similar to the ones that the ML model misclassified in order to induce better learning of the misclassified items. Or it may use a state-of-the-art active learning technique.

\section{Data Selection Algorithm}
\label{section:algorithm}

In this section we present our algorithm for selecting data for labeling. This algorithm can be used both in the \textbf{Pick more data instances} step as well as in the initial data set selection step for bootstrapping the system.

At a high level, the goal of the data selection algorithm is to identify ``good'' data instances for labeling. Since our goal is to use our resources as efficiently as possible, we need to pick data instances in a way that is the most helpful to the ML task at hand.

Intuitively, the data selection algorithm needs to select data, based on three key factors. First, at any given step of the algorithm there is a set of data $D_l$ that contains data instances that are already labeled by the humans. Based on this set, the algorithm may choose to pick more data that are similar to the existing ones, data that are under-represented in the current labeled dataset, or a mix of the two.
Second, at any given step there is a set of data instances $D_e$ that are handled erroneously by the ML model at hand. For example, if the ML model was a spam classifier, $D_e$ would contain the instances that were misclassified. The data selection algorithm may choose to pick more instances similar to those from $D_e$ (similar to the boosting and bagging techniques) in order to increase the performance of the model.   
Finally, the data selection algorithm may want to examine the set of the unlabeled data $D_u$ in order to pick good candidates for labeling. For example, the algorithm may want to pick data that are sufficiently different from previously selected data in order to have a nicely diversified dataset.
Taking these three datasets $D_l$, $D_u$ and $D_e$ into account the goal of the algorithm is to output a set $D_h \subset D_u$ of instances to be sent for labeling. Given the resource constraints that we specified in the previous section, the instances selected for labeling should not use up more than the allocated budget $R$.

Based on the ideas that we have just described, we can examine the three datasets using a variety of criteria in order to make our decision on which data instances to pick. In this paper, we focus on a method that we have found to work well in practice and is based on the idea of maximizing the coverage of the attribute-value pairs in the data instances that we have selected for labeling. More specifically, assume that the data instances in our dataset (labeled or not) are described by a set of $n$ attributes $A=\{a_1, \dots, a_n\}$ with each attribute getting values from a discrete domain, i.e. $a_i \in \{v_{i1}, \dots, v_{ik}\}, a_j \in \{v_{j1}, \dots, v{jm}\}, \dots$.\footnote{Attributes can be numerical, categorical or nominal in terms of the values they can take. The categorical and nominal values are already discretized. In the case of numerical attributes we can discretize their values in order to simplify computations. 
}
Given these assumptions, our goal is to identify data instances to label so that we can cover all the attribute-value pairs in the dataset at least $w > 1$ times. 

This approach has two main advantages. First, we will be giving the ML model a diversified dataset to train on and therefore we ensure that it will examine all of the potential space of attribute values. Second, by maximizing coverage, we hope that the ML model will be able to handle gracefully instances that were not seen verbatim during training. This is especially important given the size of the candidate training data coming from the Web, since it is nearly impossible to use all of the data for training.

The intuition behind maximizing the coverage is that in most cases the data required to solve ML problems at Web scale contain long-tail distributions in the values of their attributes. In other words there are several data instances that have a given value for a particular attribute and several data instances that individually have different values for the same attribute. As an attribute one may consider different things depending on the problem. For example, in the case of Web-page classification in two classes, sport and non-sport, one may consider the number of times that the word football appears in the content of the page as an attribute.

\begin{figure}[tpb]
  \centering
  \includegraphics[width=.7\columnwidth]{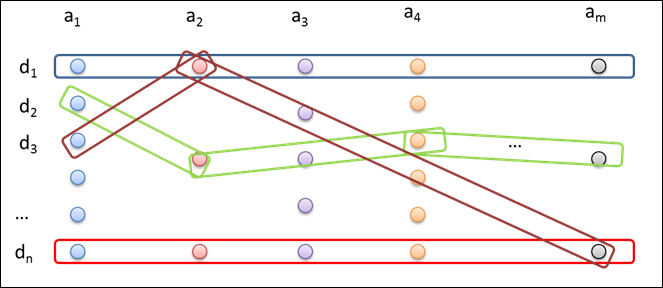}
  \caption{Attribute value coverage.}
  \label{fig2}
\end{figure}

Maximizing the coverage is not a straightforward problem. Consider the drawing shown in Figure~\ref{fig2}. Each of the columns represents an attribute and different points represent different values for that particular attribute (potentially after discretization). For each data instance we have a set over all attributes, i.e. we can see each instance as a set containing one (and only one) value from every column. Given this formalization our goal is to find the minimum number of instances that cover all possible attribute values (i.e. points) in the drawing. This is similar to the set-cover problem which is known to be NP-hard. Therefore an optimal solution cannot be found in polynomial time. 

To this end, we consider a greedy algorithm. We iteratively go over the unlabeled data instances and for each one of them we compute the number of new attribute-value pairs that they will be bringing into our set. Then we pick the instances with the highest number of new attribute-value pairs and we send it for labeling. We repeat the process until all of our resources are exhausted or we have covered all the attribute-value pairs at least $w$ times. When $w>1$ our goal is to enlarge and enrich our dataset in order to better capture potential correlations amongst attribute-value pairs.

\section{Experimentation and Evaluation Results}
\label{section:evaluation}

In this section we show experimental evidence of the performance of our system. To this end, we present experimental results on two datasets collected from the Web, namely the UCI Email data set and a dataset from Twitter. For both datasets, we considered a classification problem and we used the system that we described in the previous sections in order to label the data, train the model and evaluate it.

\subsection{UCI Email data set}

For our first experiment, we considered a spam-email dataset coming from the University of California Irvine's ML repository.\footnote{http://archive.ics.uci.edu/ml/datasets/Spambase} 
The dataset contains about 4600 emails which have been manually labeled as spam or non-spam. Each email is represented as a set of 57 attributes, which essentially represent the frequency of a set of 57 pre-selected words. Every email has a value for each of the 57 attributes (i.e. there were no missing values in our dataset). Note that by using a pre-labeled dataset, we are emulating the case where each data item would be sent to human judges to be labeled (i.e. we use the label of a data item as a proxy of what the human labeler would label the item with). 

Based on this dataset, the ML task that we considered is the training of a classifier that can discern the spam and non-spam emails based on the 57 pre-computed attribute-value pairs. Since our goal here is not to necessarily determine the best possible classifier for this task but rather investigate the performance of our system in labeling, we chose to employ a Naive Bayes classifier. Different classifiers yield slightly different performance values in the graphs but the overall trends that we report here are consistent.

We have run the dataset through our system and we present the results in Figure 3. The horizontal axes represent state-of-the art measures for measuring the performance and accuracy of a classifier, such as precision, recall, false positives, true positives and F-measure. All of the techniques were evaluated against a golden set of 900 instances that were selected randomly from the whole dataset. 

\begin{figure}[tpb]
  \centering
  \includegraphics[width=.75\columnwidth]{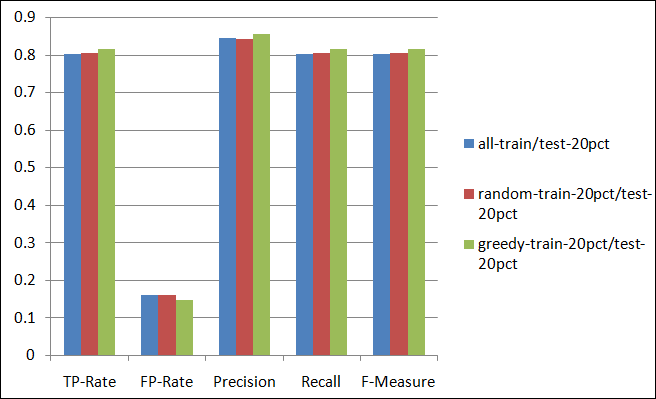}
  \caption{Performance of Naive Bayes Classification on the UCI SpamBase Dataset.}
  \label{fig3}
\end{figure}

The blue bars in Figure~\ref{fig3} correspond to the case where we use the whole dataset for training. In this case, we would spent resources proportional to the size of the whole dataset. 
To compare against different data selection algorithms, we implemented a version that randomly picked about 20\% of the data (912 instances) that were used to train. Assuming that the labeling cost is proportional to the size of the dataset, this algorithm would use only 20\% of the resources compared to the case that we used the whole dataset. We represent the performance of this algorithm using the red bars in Figure~\ref{fig3}.

Finally, we employed the data selection method described in the previous sections. In this particular case, we allowed our method even less data items to train on, more specifically, 850 instances (about 18.5\% of the dataset). For the attribute-value pairs, we discretized by rounding the number to the nearest tens-digit (e.g. 12 was rounded to 10, 27 to 30). The numbers that we report for the three data selection policies are averaged after 10-fold cross validation.  

Overall, the graphs in Figure~\ref{fig3} indicate that we can achieve performance similar to what we would have achieved if we paid the cost for labeling the whole dataset, but with much fewer resources. Specifically, our algorithm used about 18.5\% of the resources compared to labeling the whole dataset, while achieving virtually the same performance across all metrics.

\subsection{Twitter data set}

For our second experiment we collected a dataset from Twitter. In this case, we also considered a classification problem. More specifically, we are interested in building a classifier that can determine interesting from non-interesting tweets. 
This dataset contains 3423 training instances which were randomly collected over a period of five consecutive days (Monday through Friday). After each batch of tweets was collected, it was immediately uploaded to Mechanical Turk, where workers were presented with a set of tweets and asked if the content was ``interesting" or ``not interesting".   Each worker was asked to label multiple tweets, and we collected five distinct judgments for each tweet. We used the same template from a similar project \cite{alonso2010}.

\begin{figure}[tpb]
  \centering
  \includegraphics[width=.75\columnwidth]{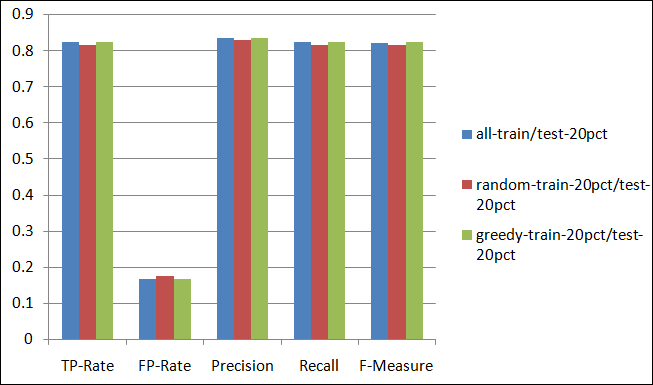}
  \caption{Performance of Naive Bayes Classification on the TwitterSpam Dataset.}
  \label{fig4}
\end{figure}

The results for the twitter dataset are shown in Figure~\ref{fig4}. Similar to the dataset presented before, we selected a gold set of 1268 instances for the evaluation of the algorithms. Again, the blue bars represent the case where we label the whole set of instances, the red bars when we randomly pick 1212 (35.4\%) of the instances and the green bars show the performance achieved by our algorithm, which selected 1000 (29.2\%) instances. As in the first dataset, our algorithm achieves similar performance compared to what we would have achieved if we had labeled the whole dataset, but uses much fewer resources (29\% of the instances).

\section{Prior Work}
\label{section:related}

How to select data for labeling falls under the subfield of machine learning known as \emph{active learning}. In active learning an algorithm selects samples for labeling by an oracle which then constitute a training set for a ML algorithm, e.g. a classifier. The goal is to minimize the number of collected labels while achieving better performance for the trained algorithm. Active learning algorithms have been considered in three main settings. Pool-based sampling methods select samples for labeling from a large pool of unlabeled data. Membership query synthesis methods can request any instance in the input space and any sample can be generated. Finally, stream-based selective methods make the assumption that the samples are being generated from a streaming process and for each sample the algorithm decides whether to select for labeling or not \cite{settles.tr09}. In most active learning algorithms, the samples are selected and labeled one at a time. While this may be an efficient method for the ML algorithm, it is very inconvenient for the human annotators, who usually would rather label big groups of data at a time. Our system operates in this manner, selecting samples in batches.

There are several approaches for selecting samples in active learning. Some algorithms select the samples for which the ML algorithm has the highest uncertainty \cite{culotta:2005,schein:2007}. Another category of algorithms use the \emph{query-by-committee} (QBC) framework. These maintain a set of competing models and try to reduce the version space, the set of model hypotheses consistent with the labeled training data. The samples that are selected come from the most controversial regions of the input space \cite{seung:1992}. Other algorithms aim to reduce the expected future error \cite{roy01}, the output variance \cite{cohn:1996}, or maximize the expected change in the model \cite{settles.emnlp08}. 

The above approaches all take into account how well the ML algorithm can model the labeled instances. This is useful when the model has already been trained on some data and can guide the selection of more instances, but the algorithms do not specify how the initial selection of training data should occur. In the majority of cases the selection is made via random sampling.
Our proposed method greedily selects samples that cover the highest number of feature values and can thus work before the ML algorithm has ever been trained. The most similar methods to our approach are density-weighted methods, which model the feature distribution of the input space, for example by clustering the instances and selecting samples that are least similar to labeled and most similar to unlabeled clusters \cite{xu:2007}. 

An important component of an active learning system is the human annotators. In the active learning literature they are considered as oracles who always provide correct and reliable labels. In practice this is never the case and significant effort must be devoted to ensure the accuracy of the returned labels.  Outside the ML community there is significant work on human computation and how to deal with the issues introduced by the human element \cite{alonso:2011}. Papers on the topic concentrate on techniques for dealing with noisy human labelers \cite{wiebe:1999}, estimating the labeler accuracy \cite{donmez:2009}, correcting label bias \cite{snow:2008} and using repeated labeling to reduce errors \cite{sheng:2008}.

Finally, practical aspects of the active learning framework that affect the number of instances that are labeled and the operation of the system are the cost of labeling and the stopping criteria. The number of labels acquired, the accuracy achieved, time spent and effort expended are all potential criteria for deciding when to stop labeling. Incorporating a labeling cost in the active learning algorithm leads to cost-sensitive active learning \cite{margineantu:2005,settles.nips08ws}. Other researchers have also included an expected future misclassification cost if an instance is not labeled \cite{kapoor:2007}. Various intrinsic criteria for stopping have also been proposed that depend on the stability of the ML algorithm (for example \cite{olsson:2009}).

\section{Conclusions and Future Work}
\label{section:conclusion}

We have presented a system for determining which data instances out of a larger dataset we should select  and send to human workers for labeling. Our system aims to reduce the cost of creating training data for ML tasks. We described the architecture of that system and the elements that comprise it. 
An important component is the data selection algorithm that determines which data instances to label. We formalized the problem as a resource optimization problem, where we are given a fixed budget $R$ and we need to determine which items to pick within this budget. Our proposed algorithm greedily selects instances that cover the most unseen feature values. Our method was evaluated on two datasets, one for classifying spam email and a second for classifying tweets as interesting or not. We found that our approach is promising in the sense that by using a fraction of the original dataset for training we can achieve performance comparable to using the whole dataset. 

An advantage of our framework is that it facilitates the identification of potential components for improvement. For example, one can try various stopping criteria, or change the way the trained models are evaluated. In future work, we plan to experiment with more datasets and different ML tasks besides classification. One question we would like to answer is how the training data that our algorithm selects affect the performance of the trained ML model and whether that effect depends on the nature of the task or the type of the ML algorithm. We also plan to experiment with alternative active learning approaches for selecting data instances, which depend on the output and performance of the ML model. Additional information about where the trained model under-performs can be used by an active learning algorithm to pick more data after the initial training iteration and could lead to better data selection, potentially further reducing the cost.

\bibliographystyle{splncs03}
\bibliography{dataset}

\end{document}